\documentclass[conference]{IEEEtran} % conference, peerreview %conference
\IEEEoverridecommandlockouts
\usepackage{amsmath,amssymb}
\usepackage{textcomp}

\usepackage{algorithm}
\usepackage{algpseudocode}
\usepackage{subcaption}
\usepackage{xspace}
\usepackage{multirow}

\usepackage[utf8]{inputenc} % allow utf-8 input
\usepackage[T1]{fontenc}    % use 8-bit T1 fonts
\usepackage{hyperref}       % hyperlinks
\usepackage{url}            % simple URL typesetting
\usepackage{booktabs}       % professional-quality tables
\usepackage{amsfonts}       % blackboard math symbols
\usepackage{nicefrac}       % compact symbols for 1/2, etc.
\usepackage{microtype}      % microtypography
\usepackage{xcolor}         % colors
\usepackage{graphicx, wrapfig}
\usepackage{titlesec}       % tile

\newcommand{\ours}{SparseBalance\xspace}
\NewDocumentCommand{\cfgMean}{o}{%
  \textsc{Mean}\IfValueT{#1}{\ensuremath{_{#1}}}\xspace
}
\NewDocumentCommand{\cfgMin}{o}{%
  \textsc{Min}\IfValueT{#1}{\ensuremath{_{#1}}}\xspace
}
\NewDocumentCommand{\cfgMax}{o}{%
  \textsc{Max}\IfValueT{#1}{\ensuremath{_{#1}}}\xspace
}
\NewDocumentCommand{\cfgAnchor}{o}{%
  \textsc{Anchor}\IfValueT{#1}{\ensuremath{_{#1}}}\xspace
}

\def\BibTeX{{\rm B\kern-.05em{\sc i\kern-.025em b}\kern-.08em
    T\kern-.1667em\lower.7ex\hbox{E}\kern-.125emX}}  
\begin{document}

\title{SparseBalance: Load-Balanced Long Context Training with Dynamic Sparse Attention}

\author{
\IEEEauthorblockN{
Hongtao Xu\textsuperscript{1,2}, 
Jianchao Tan\textsuperscript{2}, 
Yuxuan Hu\textsuperscript{2}, 
Pengju Lu\textsuperscript{1}, 
Hongyu Wang\textsuperscript{1}, 
Pingwei Sun\textsuperscript{2}, \\
Yerui Sun\textsuperscript{2}, 
Yuchen Xie\textsuperscript{2}, 
Xunliang Cai\textsuperscript{2}, 
Mingzhen Li\textsuperscript{3,*}, and 
Weile Jia\textsuperscript{3,*}
}

\vspace{0.2cm} % 同样，如果极其缺少空间可以删掉这行

\IEEEauthorblockA{
\textsuperscript{1}\textit{School of Advanced Interdisciplinary Sciences, University of Chinese Academy of Sciences}, Beijing, China \\
\textsuperscript{2}\textit{Meituan}, Beijing, China \\
\textsuperscript{3}\textit{University of Chinese Academy of Sciences}, Beijing, China
}
}

\maketitle

\begin{abstract}

While sparse attention mitigates the computational bottleneck of long-context LLM training, its distributed training process exhibits extreme heterogeneity in both \textit{1)} sequence length and \textit{2)} sparsity sensitivity, leading to a severe imbalance problem and sub-optimal model accuracy. Existing algorithms and training frameworks typically focus on single issue, failing to systematically co-optimize these two problems. 
Therefore, we propose SparseBalance, a novel algorithm-system co-design framework, which exploits the sparsity and sequence heterogeneity to optimize model accuracy and system efficiency jointly. 
First, we propose workload-aware dynamic sparsity tuning, which employs a bidirectional sparsity adjustment to eliminate stragglers and exploit inherent bubbles for free accuracy. Second, we propose a sparsity-aware batching strategy to achieve coarse-grained balance, which complements dynamic sparsity tuning.
Experimental results demonstrate that SparseBalance achieves up to a 1.33$\times$ end-to-end speedup while still improving the long-context capability by 0.46\% on the LongBench benchmark. 
\end{abstract}


\section{Introduction}
\label{sec:intro}

The long-context modeling capability is increasingly crucial for the evolution of Large Language Models (LLMs), serving as the backbone for advanced downstream applications such as code generation, in-depth reasoning, and autonomous agent systems. Mainstream LLMs typically perform an additional training stage on specific long-context datasets to extend the context windows~\cite{geminiteam2024gemini15unlockingmultimodal, glm2024chatglmfamilylargelanguage, qwen25_tech_report_2024, yang_qwen3_2025}. However, as the context length increases, the standard attention mechanism becomes the primary computational bottleneck. Because standard attention requires computing affinities across all token pairs, it exhibits a quadratic computational complexity with respect to sequence length, thus severely hindering the context expansion. To mitigate this, sparse attention has emerged as a promising solution in recent released LLMs~\cite{deepseekai2025deepseekv32pushingfrontieropen,glm5team2026glm5vibecodingagentic}. By selectively computing the critical tokens, sparse attention reduces the computational complexity and breaks the performance bottleneck. 

However, long-context sparse training still faces a severe load imbalance problem caused by the inherent heterogeneity in sequence length distributions. For instance, the Qwen2.5 technical report~\cite{qwen25_tech_report_2024} discloses their data mix strategy at long context training stage, which comprises 40\% long sequences and 60\% short sequences. This heterogeneity in sequence length leads to severe load imbalance problem in distributed training, significantly degrading the system efficiency. Existing works attempt to alleviate this issue through batching or packing strategies. However, these methods can hardly solve the problem completely due to the inherently discrete nature in this bin-packing problem, especially for highly skewed datasets. This imbalance manifests across multiple parallelism dimensions in distributed training. In inner-level pipeline parallelism (PP), varying workloads exacerbate pipeline bubbles, and the processing time of extremely long sequences essentially dominates the overall pipeline latency. Worse still, this imbalance in PP ultimately propagates to the outer-level parallelism like data parallelism (DP), severely amplifying the degradation of system efficiency.

Meanwhile, sparse training introduces another critical heterogeneity in sparsity sensitivity. Although existing trainable sparse attention algorithms typically adopt a predefined sparsity degree, \textit{\textbf{we observe distinct sparsity sensitivities across different sequences and transformer layers}}. Furthermore, if simply transitioned to dynamic sparse training without system awareness, this inherent sparsity heterogeneity would lead to the exact same workload imbalance problem. Therefore, these two dimensions of heterogeneity—sequence length and sparsity sensitivity—are deeply intertwined and jointly impact the runtime workload, highlighting the critical need for algorithm-system co-design. However, current algorithms and training frameworks typically address these challenges in isolation, solely focusing on workload balance or developing sophisticated sparse algorithms. Consequently, they fail to systematically co-optimize these two issues, resulting in sub-optimal performance in either training efficiency or model accuracy.

\textit{\textbf{ 
To address sequence length heterogeneity and sparsity sensitivity heterogeneity simultaneously, we propose \ours{}, a novel algorithm–system co-design enabling bidirectional sparsity tuning for LLM training.
\ours{} jointly optimizes model accuracy and system efficiency while considering their inherent trade-off.
}}

First, we propose a workload-aware \textbf{D}ynamic \textbf{S}parsity \textbf{T}uning (DST) strategy, which performs bidirectional sparsity adjustment at runtime to rebalance the workload at the layer level. 
Specifically, DST identifies bottleneck micro-batches and reduces their attention budget to accelerate execution, while simultaneously increasing the attention budget of non-bottleneck micro-batches to exploit pipeline bubbles for free accuracy improvements. 
To ensure the efficiency and accuracy during such tuning process, we introduce an anchor-guided thresholding mechanism, which determines the direction and bounds the magnitude for each micro-batch. 

Second, to fully unleash the optimization potential of DST, we propose a \textbf{S}parsity-\textbf{A}ware \textbf{B}atching (SAB) strategy, which involves lightweight sparsity estimation and latency-based data packing. SAB provides a well-balanced initial workload distribution, which serves as a foundation for the fine-grained runtime adjustments in DST. Additionally, we implement the latency prediction module, which maps the sequence length and sparsity to practical execution latency through offline profiling, providing the accurate performance guide for both DST and SAB module. 
Together, SAB and DST form a unified optimization pipeline, spanning from coarse-grained data reorganization to fine-grained runtime tuning, enabling efficient load balancing without sacrificing model accuracy.

We evaluate \ours{} on two real-world datasets and conduct comprehensive evaluations across three downstream benchmarks. Our key contributions are as follows:
\begin{itemize}  
    \item We provide a novel perspective on improving sparse attention training through algorithm-system co-design to jointly optimize system efficiency and model accuracy. % while explicitly managing the trade-off between system efficiency and model accuracy..
    % heterogeneity as a dynamic knob to address the system-level load imbalance problem in distributed long-context training.
    \item We propose workload-aware dynamic sparsity tuning (DST) to dynamically rebalance the training workload at runtime, while preserving model accuracy.
    \item We propose sparsity-aware batching (SAB) dedicated for sparse training scenario, it involves a lightweight sparsity estimator and latency-based batching strategy, providing coarse-grained balance for DST.
    % SAB and DST work seamlessly together to achieve optimal load balancing
    \item We implement \ours{} and experimental results demonstrate that \ours{} improves the end-to-end training efficiency by up to 1.33$\times$ while still improving the model's downstream capability.
\end{itemize}

\section{Background}  
\subsection{Distributed Training Strategies}

\textit{Data Parallelism (DP)} \cite{pytorch_distributed} partitions the global training batch across multiple workers. In each iteration, every worker independently processes its assigned subset of data and computes the local gradients. At the end of each iteration, DP enforces a rigid synchronization barrier, requiring all workers to synchronize their gradients before updating the model weights. This inherent synchronization mechanism forces all workers to wait for the slowest worker (i.e., the straggler) to reach the barrier, making DP highly sensitive to workload imbalances.

\textit{Pipeline Parallelism (PP)} \cite{gpipe_2019,pipedream_1F1B_2019,megatron_v2_3d_par_interleave1f1b_2021,zerobubble_pp_2024} splits the model layers into sequential stages, with each stage assigned to a different PP worker. To maximize hardware utilization, existing methods orchestrate model execution into a pipeline by dividing the input data into multiple micro-batches. These micro-batches then sequentially traverse all PP ranks via point-to-point communication to deliver activations and gradients between stages. However, PP inherently suffers from the pipeline bubble problem. This issue is severely exacerbated under highly heterogeneous workloads, as any execution delay in a single stage cascades throughout the entire pipeline.

\textit{Tensor Parallelism (TP)} \cite{megatron_v1_tp_2020} divides the tensor operations across devices, and each device handles a slice of the tensor operations. With TP, each GPU only has part of the input and parameters, resulting in intensive communication during training. Therefore, TP is typically applied within a single node, while other levels of parallelism are applied across nodes.

% 如何更好地解释context parallel和sequence parallel, 如何引导说明我们默认使用SP。
\textit{Sequence Parallelism (SP)} partitions the input tensor along the sequence length dimension \cite{megatron_v3_sp_selective_recompute_2023, deepspeed_ulysess_2024, ring_attention_2024, llama3_tech_report_2024}. There are three types of SP according to the communication schemes: (i) ring-based point-to-point communications \cite{ring_attention_2024}, (ii) AllToAll-based communications \cite{deepspeed_ulysess_2024}, and (iii) AllGather-based communications \cite{llama3_tech_report_2024}. Additionally, there is another type of SP, which is proposed by Megatron-LM and splits the dropout and normalization module activation \cite{megatron_v3_sp_selective_recompute_2023}. In this paper, we use this Megatron-style SP as default.

\subsection{Self-Attention and Trainable Sparse Attention}
\label{sec:bg_sparse_attention}

Modern LLMs are typically built by stacking many Transformer layers, each of which contains a self-attention module as the core mechanism to capture complex dependencies between elements within a sequence. In long-context training, the self-attention module becomes the dominant computational bottleneck because its cost scales quadratically with the sequence length. Given the query ($Q$), key ($K$), and value ($V$) matrices, where $Q, K, V \in \mathbb{R}^{N \times d}$ with $N$ denoting the sequence length and $d$ representing the hidden dimension. The self-attention computation is formulated as follows:

\begin{equation}
    \text{Attention}(Q, K, V) = \text{softmax}\left(\frac{QK^\top}{\sqrt{d}}\right)V
\end{equation}

In this formulation, the dot product $QK^\top$ calculates the raw affinity scores, representing the alignment between sequence elements. These scores are subsequently scaled by $\frac{1}{\sqrt{d}}$. Then, the $\text{softmax}$ operation is applied row-wise to yield a normalized distribution of attention weights. Finally, these weights are multiplied by the value matrix $V$ to aggregate the relevant contextual information into the final output representation.

To alleviate the quadratic computational bottleneck of standard attention in long-context modeling, various trainable sparse attention methods have been proposed \cite{lu2025mobamixtureblockattention,deepseekai2025deepseekv32pushingfrontieropen, yuan-etal-2025-nsa}. These methods leverage the inherent sparsity of the attention mechanism by selecting only a subset of highly relevant tokens, referred to as \emph{critical tokens}, to approximate the full attention computation. 

Existing trainable sparse methods typically adopt a block-sparse paradigm, which partition the key and value (KV) sequences into discrete blocks. They compute the correlation between each query token and these blocks using specific routing metrics to determine the critical tokens. This type of sparse attention computation is formulated as follows:

\begin{equation}
\label{eq:sparse_attn}
    \text{SparseAttn}(Q, K, V) = \text{softmax}\left(\frac{Q K[I]^\top}{\sqrt{d}}\right)V[I]
\end{equation}
where $I \in \mathbb{R}^{N}$ serves as the indexer indicating the set of selected critical tokens. 

The primary distinction among existing algorithms lies in the design of this routing mechanism, also referred to as the indexer. For instance, MoBA~\cite{lu2025mobamixtureblockattention} identifies critical tokens by calculating the relevance between each query and the mean representation of the key blocks. Similarly, DSA~\cite{deepseekai2025deepseekv32pushingfrontieropen} employs a lightweight, low-precision indexer based on multi-head latent attention. Despite differing indexer designs, these block-sparse algorithms uniformly select a fixed Top-$K$ most relevant token blocks for computation across all samples. Notably, our method reuses the inherent indexer of these sparse algorithms to achieve workload-aware dynamic sparsity tuning with negligible additional overhead. Consequently, our approach can be seamlessly integrated into existing sparse attention algorithms.
In this paper, we also use the \emph{attention budget} $k$ (e.g., the Top-$K$ selected blocks) to characterize the sparsity level. A smaller attention budget $k$ indicates higher sparsity, while a larger $k$ indicates lower sparsity.

\section{Observation}
\label{sec:motivation} 
\subsection{Sequence Length Heterogeneity in Datasets}

\begin{figure}
    \centering
    \hspace*{-0.11\linewidth}
    \includegraphics[width=0.9\linewidth]{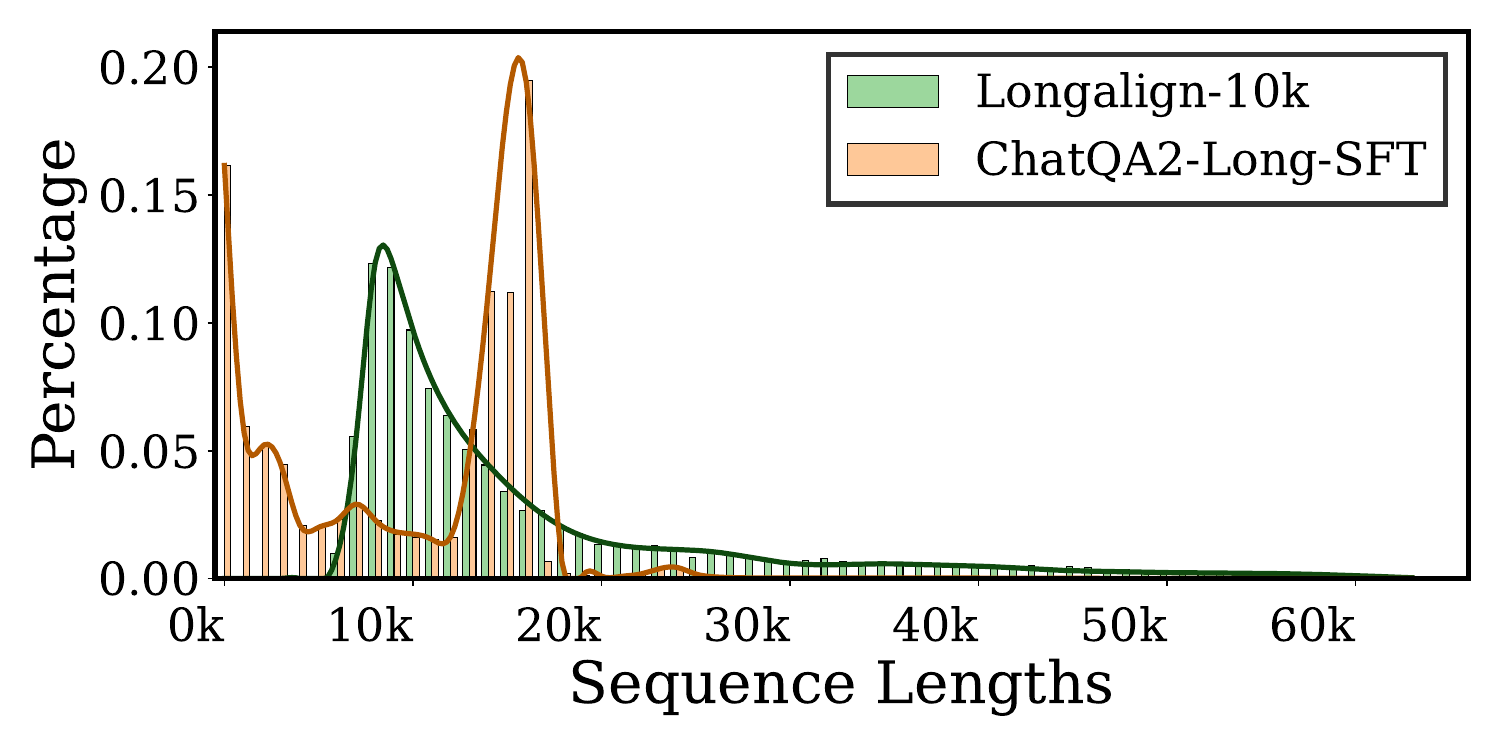}
    \caption{Extreme heterogeneity in sequence length distributions across two real-world long-context datasets. The x-axis represents the sequence length, and the y-axis denotes its corresponding proportion within the entire dataset. }
    % The yellow and green areas illustrate the distributions for ChatQA2-Long-SFT and LongAlign-10k, respectively.}
    \label{fig:varying_length}
\end{figure}

In long context training, the sequence length distribution could varying significantly. To illustrate this, we analyze two real-world, open-source datasets: nvidia/ChatQA2-Long-SFT-data (ChatQA2) \cite{dataset_chatqa2-long-sft-data_2025} and zai-org/LongAlign-10k \cite{bai-etal-2024-longalign}. As shown in Figure~\ref{fig:varying_length}, both datasets exhibit highly skewed length distributions. ChatQA2 demonstrates a bimodal pattern, dominated by ultra-short (<4K) and long (>16K) sequences. In contrast, LongAlign-10k presents a pronounced long-tail distribution, with sequence lengths stretching continuously from 8K up to 72K tokens. Consequently, this extreme heterogeneity in sequence lengths leads to distinct runtime workloads across distributed workers, highlighting the critical need for a workload-aware solution that can dynamically balance workloads.

\subsection{Imbalance Problem in Distributed Training}
\label{sec:obs_imbalance_analysis}
\begin{figure}
    \centering
    \includegraphics[width=1\linewidth]{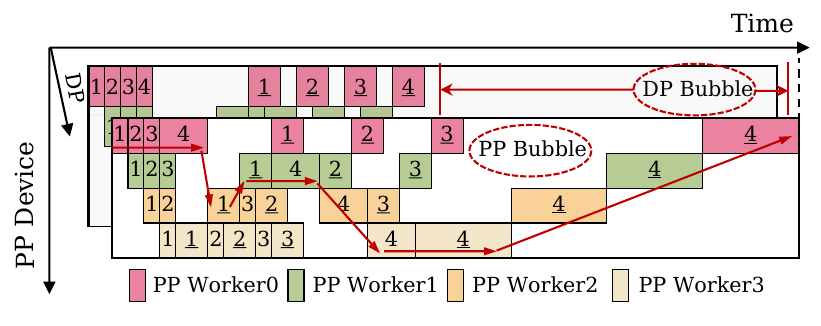} 
    \caption{Illustration of the straggler effect caused by workload imbalance. Within PP group, a bottleneck micro-batch (e.g., Micro-batch 4) dictates the critical path and exacerbates pipeline bubbles. Furthermore, DP synchronization significantly amplifies this imbalance across the entire system.}
    \label{fig:imbalance_analysis}
\end{figure}

\begin{figure}
    \centering
    \includegraphics[width=1\linewidth]{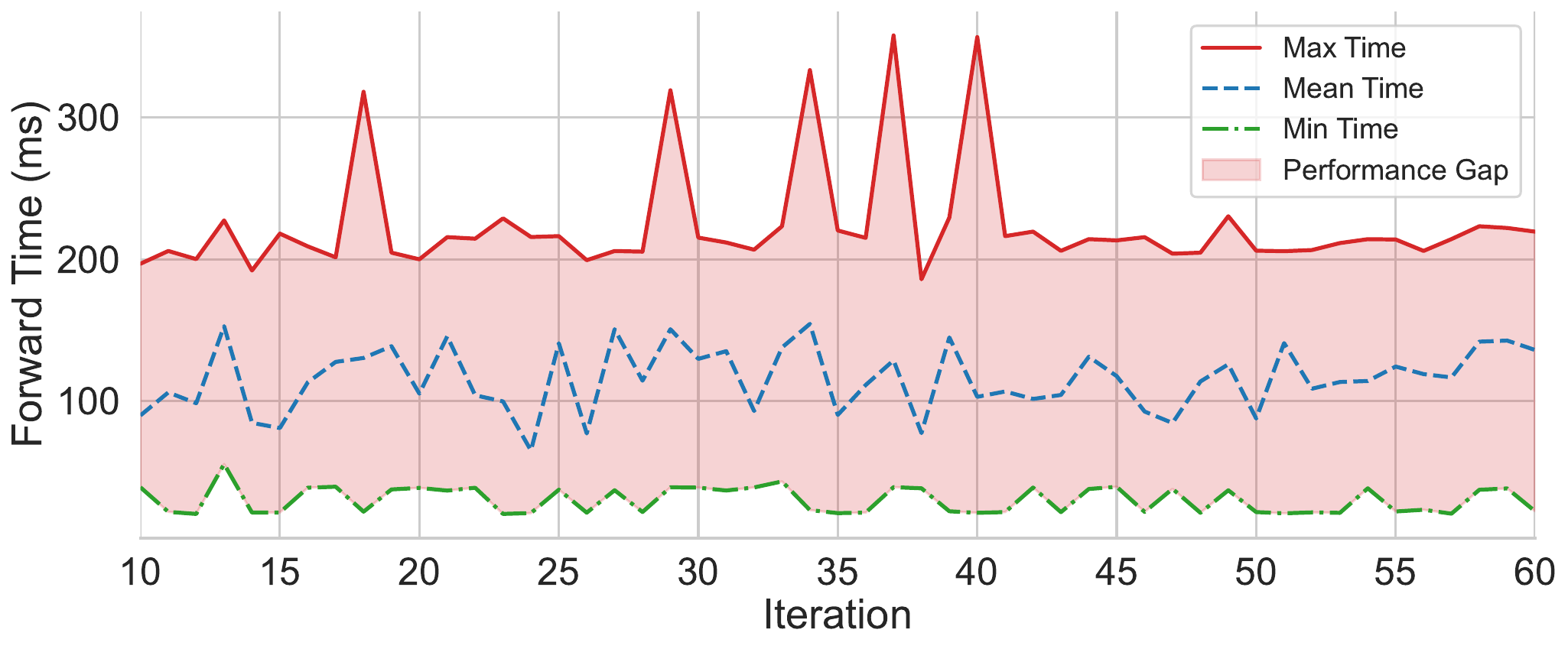}
    % \caption{Imbalance execution time across rank and iteration.}
    \caption{Forward-pass latency of micro-batches over continuous training iterations. The substantial gap between the maximum ($T_{max}$, red solid line), mean ($T_{mean}$, blue dashed line), and minimum ($T_{min}$, green dashed line) latencies reveals severe workload imbalance.} 
    \label{fig:imbalance_time}
\end{figure}

The heterogeneous sequence length distributions directly degrade the distributed LLM training, leading to severe load imbalance problem. The parallel training strategies always operate in a hierarchical and synchronized manner. The longest micro-batch inevitably bottlenecks the entire system. From the perspective of pipeline parallelism (PP), stages process micro-batches sequentially and rely on synchronous point-to-point communication to pass middle states. As illustrated in Figure~\ref{fig:imbalance_analysis}, the critical path of the PP execution time can be formulated as $T_{pp} \geq T_{max} \times PP\_Size + \sum{T_{other}}$
% revise the variable such as N_{pp}
. Therefore, the longest micro-batch strictly bounds the overall pipeline throughput, causing severe pipeline bubbles. Data parallelism (DP) also suffers from the straggler effect due to the requirement of collective gradient synchronization at the end of each iteration. Worse still, the imbalance will be amplified when DP and PP are coupled, which is a standard configuration in large-scale LLM training. Consequently, the longest micro-batch fundamentally dominates the global step time, severely impacting the end-to-end training latency.

To empirically quantify this phenomenon, Figure~\ref{fig:imbalance_time} visualizes the forward-pass latency of various micro-batches over 50 continuous training iterations. Specifically, the red solid line represents the maximum latency ($T_{max}$), the blue dashed line indicates the mean latency ($T_{mean}$), and the green dashed line denotes the minimum latency ($T_{min}$). The substantial performance gap between the maximum and minimum latencies indicates a massive optimization potential. While existing works resort to more balanced batching strategies, they cannot perfectly eliminate the straggler effect due to the discrete nature of the batching problem, especially for highly skewed data distributions, highlighting the critical need for a more fundamental approach. 

\subsection{Inter- and Intra-Sequence Sparsity Heterogeneity}
\label{sec:sparse_heterogeneity}
\begin{figure}
    \centering
    % \hspace*{-0.08\linewidth}
    \includegraphics[width=0.8\linewidth]{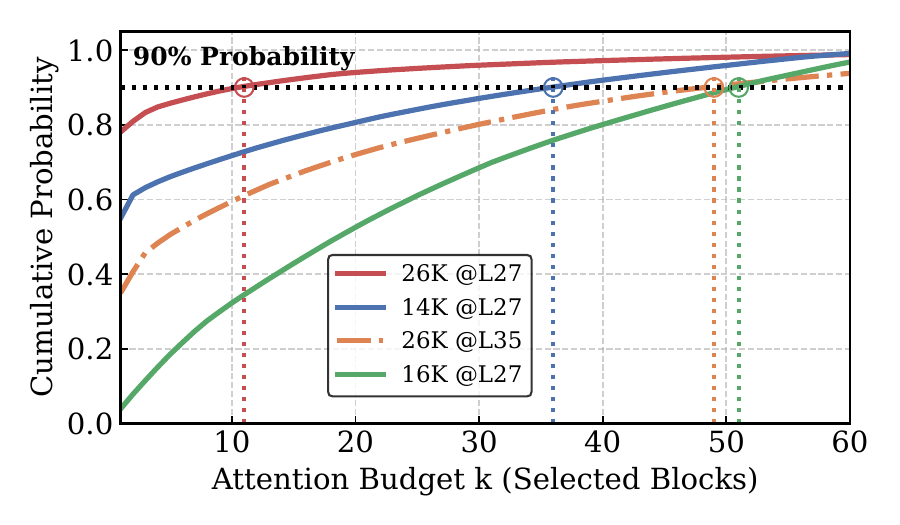}
    \caption{Cumulative score coverage under different attention budgets.}  
    \label{fig:sparse_heterogeneity}
\end{figure}

Beyond sequence length heterogeneity, sparse training also exhibits substantial \textbf{\emph{heterogeneity in sparsity sensitivity}}, at both the inter-sequence and intra-sequence levels. To quantify this effect, we profile the cumulative probability of attention score in our realistic training scenario. Figure~\ref{fig:sparse_heterogeneity} plots the recovery ratio under different attention budgets $k$ (equivalently, the number of selected critical blocks (TopK), more budgets implies less sparsity degree). We report three sequences of lengths 14K, 16K, and 26K at layer 27, and additionally plot the same 26K sequence at layers 27 and 35.

First, The results reveal clearly distinct sparsity sensitivities across sequences, which we refer to \textbf{\emph{inter-sequence sparsity heterogeneity}}. For instance, the red curve in Figure~\ref{fig:sparse_heterogeneity} exhibits a highly concentrated attention pattern, where a minimal budget (about $ k \approx 10$) is sufficient to recover 90\% of the original attention representation. By contrast, the blue and green curves demonstrate a dispersed attention distribution, requiring a substantially larger budgets (about $ k\approx 36$ and $50$) to achieve the same recovery ratio. 

Second, the same sequence can exhibit very different sparsity patterns across layers, which we term \textbf{\emph{intra-sequence sparsity heterogeneity}}. For the same 26K sample shown in Figure~\ref{fig:sparse_heterogeneity}, layer 27 (red) remains sharply concentrated, whereas layer 35 (orange) is much flatter and more sensitive to aggressive budget reduction. 

Taken together, these observations reveal substantial potential for dynamic sparsity tuning and directly motivates the layer-level dynamic sparsity tuning design in \ours.

\section{Design} 

\begin{figure}
    \centering
    \includegraphics[width=1\linewidth]{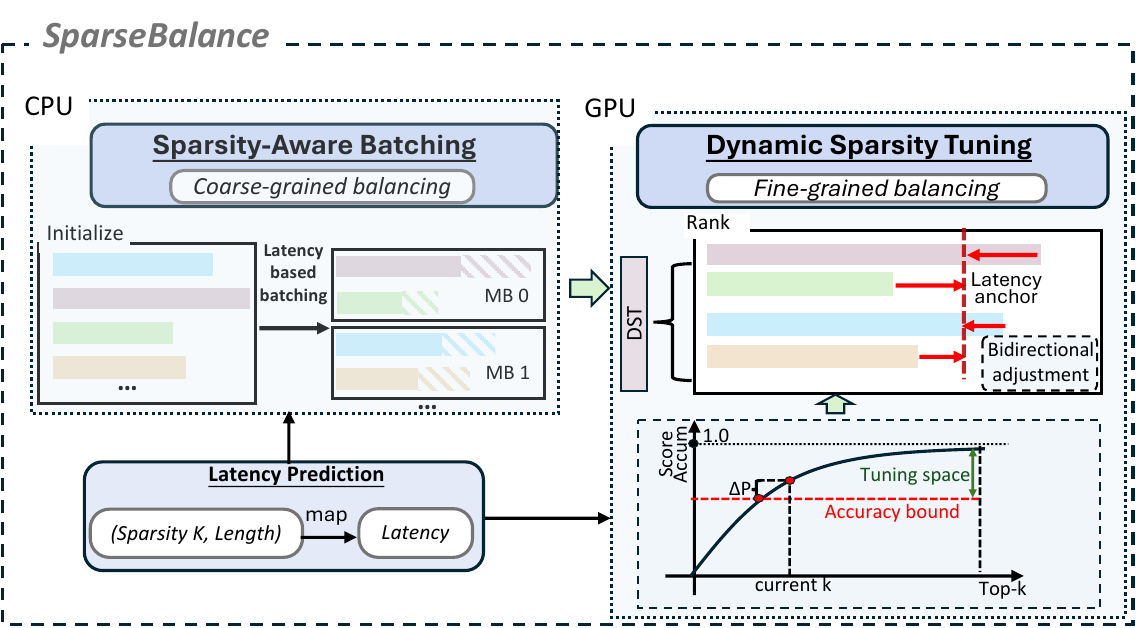}
    \caption{Overview of \ours{}, which consists of three components: workload-aware dynamic sparsity tuning (DST), sparsity-aware batching (SAB) and latency prediction module.}
    \label{fig:overall}
\end{figure}

Figure~\ref{fig:overall} illustrates the overall design of \ours{}. From the execution perspective, \ours{} balances the workload in a coarse-to-fine manner. SAB first reorganizes training samples using sparsity estimation and latency-based packing, achieving coarse-grained workload balance across micro-batches. Built on top of this, DST further conducts per-layer runtime sparsity tuning to mitigate residual imbalance at a finer granularity while preserving model quality. Both SAB and DST rely on accurate latency prediction as a performance guidance for optimization. To this end, we design a latency prediction module that maps sequence length and sparsity degree to practical execution latency under specific hardware and training configurations through offline profiling manner. For clarity, we first present DST in Section~\ref{sec:DST}, then introduce the latency prediction module in Section~\ref{sec:offline_profiling}, and finally describe the design of SAB in Section~\ref{sec:SAB}.

\subsection{Workload-Aware Dynamic Sparsity Tuning}
\label{sec:DST}

Existing methods for long-context training typically rely on batching or packing strategies to alleviate the imbalance problem induced by the heterogeneity in sequence length. 
However, due to the inherently discrete nature of batch construction, these approaches cannot fundamentally eliminate the imbalance, especially under highly skewed sequence-length distributions. Even with careful data shuffling, a single micro-batch containing a difficult-to-pack sequence may still become a straggler on the critical path, delaying the entire iteration. Therefore, a finer-grained load balancing mechanism is required to mitigate this problem.

Fortunately, sparse training provides such an opportunity for finer-grained load balancing. As discussed in Section~\ref{sec:sparse_heterogeneity}, long-context sparse training exhibits another important heterogeneity in sparsity sensitivity. This observation allows us to treat the sparsity (attention budget) not merely as a fixed algorithmic hyper-parameter, but as an adaptive degree of freedom for runtime load balancing. Based on this insight, we propose \textit{\textbf{Workload-Aware Dynamic Sparsity Tuning (DST)}}, which dynamically redistributes the attention budget across micro-batches according to their impact on the critical path.

At a high level, DST performs \emph{bidirectional} budget adjustment. For bottleneck micro-batches on the critical path, DST reduces their attention budget, i.e., tunes them toward higher sparsity, to shorten their execution time and eliminate stragglers. Even a small reduction for these bottlenecks can improve the end-to-end step latency, since their delay is amplified by synchronization and dependencies in distributed training. In contrast, for non-bottleneck micro-batches whose latency does not affect the overall step time, DST allocates more attention budget, i.e., tunes them toward lower sparsity, to improve the fidelity of sparse attention. Although such adjustment may slightly increase local execution time, the additional cost can be absorbed by the inevitable idle time introduced by pipeline execution and synchronization. In this sense, DST converts this otherwise wasted time into \textit{free accuracy}.

To make this bidirectional tuning both effective and safe, we design an \textit{\textbf{anchor-guided thresholding mechanism}}. For a global batch containing $N$ micro-batches, let $k_i^{base}$ denote the attention budget assigned by the underlying sparse attention method. Using the latency predictor introduced in Section~\ref{sec:offline_profiling}, we first estimate the latency of each micro-batch under its base budget, and then define an execution anchor as
\begin{equation}
T_{\text{anchor}}
=
\phi \!\left(
\left\{
\hat{T}_i(k_i^{base})
\right\}_{i=1}^{N}
\right)
\label{eq:anchor}
\end{equation}
where $\phi(\cdot)$ is an anchor operator, such as the minimum, mean, or maximum of the predicted micro-batch latencies. DST then aligns each micro-batch to this reference point and obtains the corresponding target budget $k_i^{anchor}$. Therefore, the anchor determines the tuning direction: micro-batches above the anchor are compressed, whereas those below the anchor can use more attention budget.

To preserve model quality, DST further constrains the tuning magnitude using the routing logits already produced by the sparse attention indexer, introducing almost no additional runtime overhead. Specifically, we normalize and sort the routing logits of micro-batch $i$ to obtain importance scores $r_{i,1} \ge r_{i,2} \ge \cdots$, where $\sum_j r_{i,j}=1$. We then define the cumulative score coverage under budget $k$ as
\begin{equation}
C_i(k) = \sum_{j=1}^{k} r_{i,j}
\label{eq:coverage}
\end{equation}
$C_i(k)$ characterizes how well the selected sparse pattern preserves the high-importance regions indicated by the routing scores. For bottleneck micro-batches, reducing the budget from $k_i^{base}$ to $k_i^{anchor}$ may hurt model quality. We therefore bound the allowable coverage drop by a threshold $p$:
\begin{equation}
C_i(k_i^{base}) - C_i(k_i^{anchor}) \le p
\label{eq:threshold}
\end{equation}
Here, $p$ serves as the primary knob controlling the aggressiveness of dynamic sparsity tuning: a larger $p$ permits more aggressive compression of bottleneck micro-batches and thus favors system efficiency, while a smaller $p$ enforces stricter fidelity to the original sparse attention pattern.

For bottleneck micro-batches, DST further restricts the candidate budgets to those that simultaneously satisfy the latency target and the quality bound. We define the feasible set as
\begin{equation}
\label{eq:feasible_set}
\mathcal{F}_i =
\left\{
k \in \mathcal{K}
\;\middle|\;
\hat{T}_i(k) \le T_{\text{anchor}},
\;
C_i(k_i^{base}) - C_i(k) \le p
\right\}
\end{equation}
Accordingly, the final accepted budget is given by
\begin{equation}
\label{eq:final_budget}
k_i^{final} =
\begin{cases}
\max \mathcal{F}_i,
& \hat{T}_i(k_i^{base}) > T_{\text{anchor}} \\[4pt]
k_i^{anchor},
& \hat{T}_i(k_i^{base}) \le T_{\text{anchor}}
\end{cases}
\end{equation}

In this way, the anchor determines the tuning direction and target balance point, while the threshold $p$ controls the safe tuning magnitude. As a result, DST shortens the critical path by compressing bottleneck micro-batches and simultaneously improves the computation fidelity of non-bottleneck micro-batches, achieving better workload balance without sacrificing model quality.

% \subsection{Latency Prediction}
\subsection{Profiling-Based Latency Modeling}
\label{sec:offline_profiling}

The effectiveness of DST relies on accurate latency guidance. In particular, DST must estimate not only the execution latency of a micro-batch under its current attention budget, but also the budget required to align its latency to a target anchor. Therefore, we design a lightweight latency prediction module that maps sequence length and attention budget to practical execution latency under a given hardware environment, parallelization strategy, and sparse attention implementation.

\begin{figure}
    \centering
    \includegraphics[width=1\linewidth]{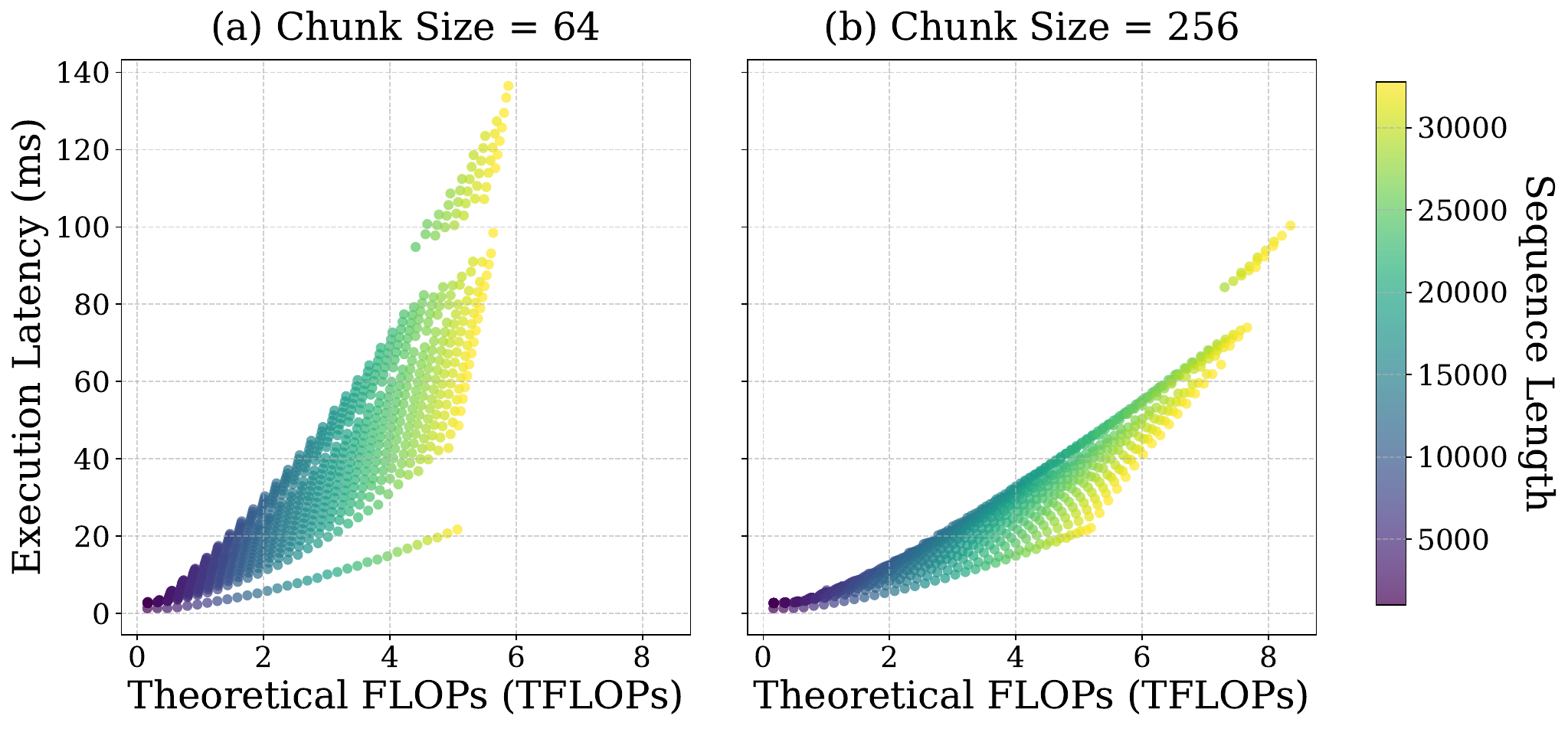}
    \caption{Practical latency of sparse attention cannot be reliably predicted by either theoretical FLOPs or sequence length alone, motivating our profiling-based latency predictor.}
    \label{fig:profiling_data}
\end{figure}

However, building an accurate latency predictor for sparse attention is non-trivial. For standard attention, sequence length or theoretical FLOPs is often a reasonable proxy for workload. In sparse training, however, this approximation becomes unreliable, because the practical latency is jointly affected by sequence length, attention budget, sparse kernel implementation, and hardware characteristics. To empirically validate this mismatch, we profile the latency of one Transformer layer under different settings using the official implementation of MoBA. As shown in Figure~\ref{fig:profiling_data}, neither FLOPs-based nor length-based modeling can consistently predict the practical latency of sparse attention. Therefore, we adopt a profiling-based design instead of an analytical one.

The predictor \texttt{Perf} exposes two runtime interfaces:
\begin{equation}
\hat{T}(x, k) = \texttt{Perf.predict}(x, k)
\label{eq:perf_predict}
\end{equation}
\begin{equation}
k^{\text{target}} = \texttt{Perf.align}(x, T_{\text{target}})
\label{eq:perf_align}
\end{equation}
where $x$ denotes the length descriptor of a micro-batch and $k$ denotes the attention budget. The first interface estimates the practical latency under budget $k$, while the second returns the largest budget whose predicted latency does not exceed the target latency $T_{\text{target}}$.

To materialize these interfaces efficiently, we adopt a \textit{\textbf{profiling-based latency modeling}} approach. Specifically, we use a single Transformer layer as a performance proxy, discretize sequence lengths into bins $\mathcal{X}$ and attention budgets into a small candidate set $\mathcal{K}$, and profile the latency of every pair $(x,k)\in\mathcal{X}\times\mathcal{K}$. The resulting lookup table is
\begin{equation}
\mathcal{M}[x,k] = \texttt{profile}(x,k)
\label{eq:lookup_table}
\end{equation}
where $\mathcal{M}$ stores the measured latency on the target system. At runtime, \texttt{Perf.predict} queries $\mathcal{M}$ directly, optionally with interpolation when the input falls between neighboring buckets, while \texttt{Perf.align} searches over $\mathcal{K}$ and returns the budget whose predicted latency best matches the target.

This design keeps the online overhead negligible, because runtime only requires lightweight table lookup and search over a small candidate set. Moreover, the profiled table remains reusable as long as the hardware environment, parallelization strategy, and sparse attention implementation remain unchanged. Based on this predictor, Algorithm~\ref{alg:DST} summarizes the runtime workflow of DST. In addition, we later reuse the same latency predictor in SAB to guide latency-based batching.

% 这个texttt大小很奇怪
\begin{algorithm}[htbp!]
% \footnotesize
\caption{Workload-Aware Dynamic Sparsity Tuning (DST)}
\label{alg:DST}
\begin{algorithmic}[1]
\Require Threshold $p$, anchor strategy $anchor$, current micro-batch length $x_i$, global-batch length list $\{x_j\}_{j=1}^{N}$, base budget list $\{k_j^{base}\}_{j=1}^{N}$, routing logits $logits_i$, latency predictor $Perf$
\Ensure Final attention budget $k_i^{final}$ for current micro-batch

\State $T_{list} \gets [~]$
\For{each $(x_j, k_j^{base})$ in $\{(x_j, k_j^{base})\}_{j=1}^{N}$}
    \State $T_{list}.\texttt{append}(\texttt{Perf.predict}(x_j, k_j^{base}))$
\EndFor

\State $T_{\text{anchor}} \gets \texttt{SelectAnchor}(T_{list}, anchor)$
\State $k_i^{anchor} \gets \texttt{Perf.align}(x_i, T_{\text{anchor}})$
\State $k_i^{final} \gets k_i^{anchor}$

\State $C_i \gets \texttt{AccumulateNormScores}(logits_i)$

\If{$C_i[k_i^{base}] - C_i[k_i^{anchor}] > p$}
    \State $k_i^{final} \gets \texttt{FindFeasibleK}(C_i, k_i^{base},p)$
\EndIf

\State \Return $k_i^{final}$
\end{algorithmic}
\end{algorithm}

\subsection{Sparsity-Aware Batching}
\label{sec:SAB}

While DST provides fine-grained runtime balancing, it still benefits from a reasonably balanced initial workload distribution. Workload balanced batching strategy can provide this initial balance and enable smoother runtime sparsity tuning in DST. However, existing batching or packing strategies typically rely on sequence length or theoretical FLOPs as the workload partition metric \cite{xu2025skrull, wang2025wlb}, which is often inaccurate in sparse attention scenario as discussed in Section~\ref{sec:offline_profiling}. Moreover, batching is executed on the CPU side before the GPU-side DST module, while the final sparsity is determined dynamically in the DST module. Therefore, the accurate workload estimation remains difficult.

To address the above issues, we implement \textit{\textbf{Sparsity-Aware Batching}} (SAB), a batching strategy tailored to dynamic sparse training, which involves lightweight sparsity prediction and latency-based workload partition. 
First, we use the base sparsity as an initial estimate and periodically calibrate it using runtime statistics collected from DST. 
Concretely, for each sequence-length bin, we maintain an exponential moving average of the final attention budgets produced by DST, and periodically refresh the sparsity estimate used by SAB. 
Although this approximation is lightweight, it is sufficient to provide a more workload-aware batching signal in practice.

Based on the predicted sparsity, we reuse \texttt{Perf.predict} as a sample-level workload proxy to guide the batching process.
% , which approximates each sample’s contribution to the eventual micro-batch latency.
As shown in Algorithm~\ref{alg:sparsity_aware_batching}, SAB first estimates the workload weight of each sample through sparsity prediction and latency prediction. 
On top of that, SAB implements a two-level workload partition, which first balances the workload across DP groups and then forms micro-batches within each group. 
Here, \texttt{bin\_packing} denotes a standard balanced partition solver that groups samples according to the provided workload weights. The output of SAB is a set of micro-batch index lists, which are then consumed by the dataloader to materialize the actual training batches and transfer to GPU.
Since this reorganization is strictly confined within the global batch, it preserves original data randomness.
% and keeps the training procedure equivalent to the original one. 
In this way, SAB provides a better coarse-grained initialization for DST, while DST further refines the residual imbalance at runtime.

\begin{algorithm}[htbp!]
\caption{Sparsity-Aware Batching (SAB)}
\label{alg:sparsity_aware_batching}
\begin{algorithmic}[1]
\Require Input batch $B$, global batch size $gbs$, micro-batch size $mbs$, data parallel size $N$, current DP rank $dp_i$, sequence lengths $L$, sparsity predictor $Spar$, latency predictor $Perf$
\Ensure Local micro-batch index bins for the current DP rank

\State $W \gets \emptyset$
\For{each sample index $i \in B$}
    \State $\tilde{k}_i \gets \texttt{Spar.predict}(L[i])$
    \State $W[i] \gets \texttt{Perf.predict}(L[i], \tilde{k}_i)$
\EndFor

\State $DP\_Bins \gets \texttt{bin\_packing}(W, N)$
% \State \Comment{$DP\_Bins[dp]$ is the sample-index set assigned to DP rank $dp$}

\State $I_{dp} \gets DP\_Bins[dp_i]$
\State $MB\_Bins \gets \texttt{bin\_packing}(W[I_{dp}],\, gbs/(mbs \cdot N))$

\State \Return $MB\_Bins$
\end{algorithmic}
\end{algorithm}

\section{Experiments}
\begin{figure*}[t]
    \centering
    \includegraphics[width=1\linewidth]{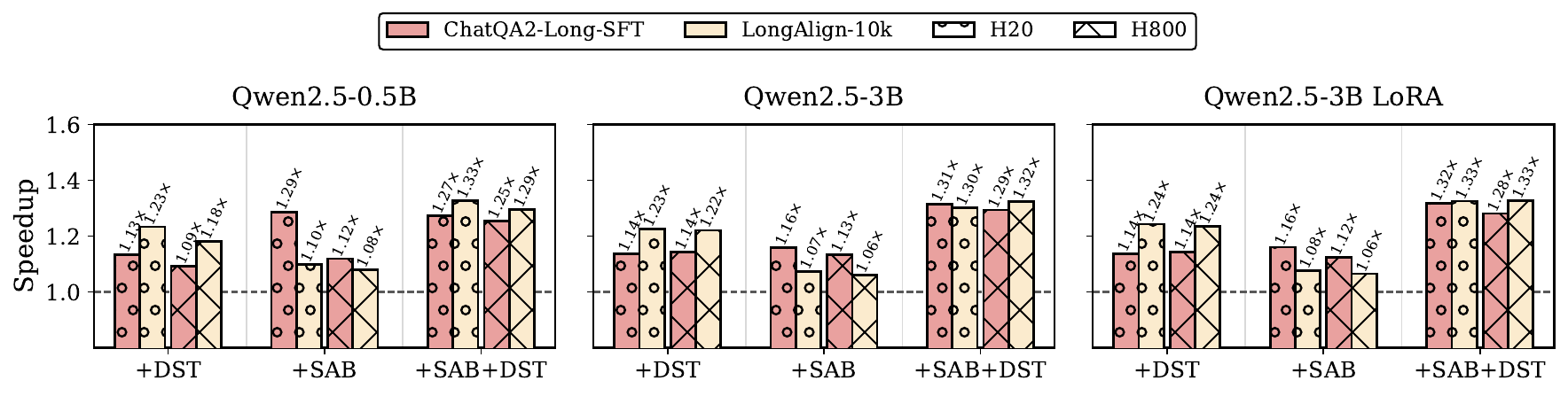}
    \caption{Overall performance and step-by-step speedups on two clusters and two datasets. We evaluate the normalized speedup ratios (measured by average iteration time) of \ours and its two main components for specific model settings (from left to right: Qwen2.5-0.5B, Qwen2.5-3B, and Qwen2.5-3B in LoRA). Results demonstrate significant efficiency improvements in \ours and confirm the effectiveness of each component.}
    \label{fig:e2e_speedup}
\end{figure*}

\begin{table*}[htbp!]
\centering
\caption{Evaluation results on the LongBench benchmark. The best results for each dataset are highlighted in \textbf{bold}.}
\label{tab:3b_eval_longbench}
% \begin{tabular*}{\linewidth}{@{\extracolsep{\fill}}llccccc@{}}
\begin{tabular}{llccccc}
\toprule
\textbf{Category} & \textbf{Dataset} & \textbf{Baseline SFT} & \textbf{\cfgMin[0.1]} & \textbf{\cfgMin[0.2]} & \textbf{\cfgMean[0.1]} & \textbf{\cfgMean[0.2]} \\
\midrule
\multirow{3}{*}{Single-Doc QA} 
& NarrativeQA & 22.42 & \textbf{23.62} & 22.88 & 22.89 & 22.03 \\
& Qasper & 32.25 & 35.77 & \textbf{36.62} & 32.99 & 34.88 \\
& MultiFieldQA-en & 42.06 & 42.19 & 41.54 & \textbf{42.92} & 42.06 \\
\midrule
\multirow{3}{*}{Multi-Doc QA} 
& HotpotQA & 44.81 & 46.80 & 45.65 & 45.40 & \textbf{47.25} \\
& 2WikiMQA & 33.42 & 33.17 & 33.98 & \textbf{35.62} & 33.20 \\
& MuSiQue & 24.30 & \textbf{26.71} & 25.50 & 26.51 & 25.78 \\
\midrule
\multirow{3}{*}{Summarization} 
& GovReport & \textbf{31.94} & 27.75 & 28.02 & 30.94 & 27.60 \\
& QMSum & 20.99 & 21.01 & \textbf{21.68} & 20.92 & 20.75 \\
& MultiNews & \textbf{25.49} & 25.38 & 25.06 & 25.04 & 24.78 \\
\midrule
\multirow{2}{*}{Retrieval} 
& TriviaQA & 86.65 & 87.21 & 87.61 & 87.67 & \textbf{88.01} \\
& PassageRetrieval-en & \textbf{11.50} & 7.00 & 9.00 & 9.00 & 10.00 \\
\midrule
\multirow{2}{*}{Code} 
& LCC & \textbf{69.82} & 68.93 & 69.45 & 69.53 & 69.19 \\
& RepoBench-p & \textbf{65.01} & 63.95 & 63.59 & 63.54 & 63.30 \\
\midrule
\textbf{Overall} & \textbf{Average} & 39.28 & 39.19 & 39.28 & \textbf{39.46} & 39.14 \\
\bottomrule
\end{tabular}
% \end{tabular*}
\end{table*}

In this section, we comprehensively evaluate \ours in both aspects of system efficiency and model accuracy. We first detail our experimental setup in Section~\ref{sec:exp_settings}. Then, Section~\ref{sec:efficiency_eval} evaluates the system efficiency brought by \ours. Following this, we analyze its impact on model accuracy and downstream benchmark in Section~\ref{sec:accuracy_eval}. Finally, Section~\ref{sec:trade_off_analysis} discusses the trade-offs between system efficiency and model accuracy. 

\subsection{Experimental Settings} 
\label{sec:exp_settings}

\textbf{Testbeds.} The experiments were conducted on two GPU clusters. The first cluster contains 4 nodes and each node equipped with 8$\times$H800 GPUs. The second cluster contains 4 nodes and each node equipped with 8$\times$H20 GPUs. Both clusters are interconnected via NVLink within the node and InfiniBand network across the node. The software stack includes CUDA 12.4~\cite{cuda2026}, PyTorch 2.5.1~\cite{PyTorch2019}, and NCCL 2.21.5~\cite{nccl2026}. We implement \ours on top of Megatron-LM~\cite{megatron_v1_tp_2020, megatron_v2_3d_par_interleave1f1b_2021, megatron_v3_sp_selective_recompute_2023} (core v0.13.2) and ms-swift~\cite{swift} (v3.10.0).

\textbf{Models, Datasets, and Baseline.} We evaluate \ours using two open-source LLMs: Qwen2.5-0.5B and Qwen2.5-3B~\cite{qwen25_tech_report_2024}. We use two real-world long-context datasets: ChatQA2-Long-SFT~\cite{xu2024chatqa} and LongAlign-10k~\cite{bai-etal-2024-longalign}, whose sequence length distributions are shown in Figure~\ref{fig:varying_length}. 
To evaluate the model accuracy, we conduct long context supervised training using Qwen2.5-3B model on ChatQA2-Long-SFT dataset with 3 epoch, selecting MoBA~\cite{lu2025mobamixtureblockattention} as our base sparse attention algorithm. To assess the long context capabilities, we evaluate the model on three downstream tasks.

\textbf{Hyper-parameters Configuration.} We select three representative anchor strategies: Mean-Anchor (\cfgMean), Min-Anchor (\cfgMin), and Max-Anchor (\cfgMax), which align the workload to the mean, minimum, and maximum of predicted micro-batch latency, respectively. For the threshold $p$, we mainly evaluate two settings, $p=0.1$ and $p=0.2$, denoted as \cfgAnchor[p]. By default, we use a hybrid 4D parallel configuration with $\text{DP}=4$, $\text{PP}=4$, $\text{TP}=2$, and $\text{SP}=2$, together with selective activation recomputation, a micro-batch size of 1, and a global batch size of 16. We set the peak learning rate to $1\times10^{-6}$, the warm-up fraction to 0.20, and the minimum learning rate to $1\times10^{-7}$. For Parameter-Efficient Fine-Tuning (PEFT) experiments with LoRA~\cite{hu_lora_2022,chen_longlora_2024}, we use rank 32, alpha 64, and dropout 0.05.

\subsection{Efficiency Evaluation}
\label{sec:efficiency_eval}
In this section, we focus on the system efficiency of \ours. We first report the overall speedups under different settings and hardware in Section~\ref{sec:e2e_perf}. Then, we conduct quantitative analysis of the workload balance situation and overheads of \ours in case study Section~\ref{sec:case_study}. Finally, in Section~\ref{sec:ablation}, we analyze the efficiency impact of the anchor selection, threshold value and micro-batch size. In the following sections, we use \cfgMean[0.1] as default configuration, as it provides the optimal overall trade-off between efficiency and model accuracy, which we detailed in Section~\ref{sec:trade_off_analysis}.

% Among them, \cfgMean[0.1] serves as our default practical configuration, as it provides the best overall balance between efficiency and model quality in our experiments.

\subsubsection{End-to-End Performance}
\label{sec:e2e_perf}

We evaluate the end-to-end performance of \ours on the two datasets and two GPU clusters described in Section~\ref{sec:exp_settings}.
We assess \ours using three model configurations: 0.5B, 3B, and 3B with LoRA. As shown in Figure~\ref{fig:e2e_speedup}, \ours achieves an average end-to-end speedup of 1.30$\times$, peaking on the Qwen2.5-3B model trained on the LongAlign-10k dataset. We observe similar speedup trends on both the H800 and H20 clusters, which suggests that the efficiency benefit of \ours is robust across different GPU platforms.
Furthermore, we perform an ablation study to isolate the effectiveness of each component: ``+DST'' denotes the exclusive application of the Dynamic Sparsity Tuning module, while ``+SAB'' indicates the standalone use of the Sparsity-Aware Batching module.

First, we analyze the performance variances between the two datasets. In terms of end-to-end speedup, LongAlign achieves higher overall gains compared to ChatQA2-Long-SFT. However, when breaking down the contributions, LongAlign benefits significantly more from the DST module, whereas its gains from the SAB module are noticeably lower. We attribute this discrepancy to the distinct length distributions and intrinsic characteristics of the datasets. As discussed in Section~\ref{sec:motivation}, sequence lengths in LongAlign are relatively concentrated between 8K and 16K tokens. Conversely, ChatQA2-Long-SFT exhibits a clear bimodal distribution, with the majority of sequences being either shorter than 4K or longer than 16K tokens. Because of this concentrated distribution, LongAlign is less sensitive to batching strategies. In contrast, the extreme length variance in ChatQA2-Long-SFT allows the SAB module to deliver larger optimizations. Furthermore, as the optimization potential of the DST module is highly correlated with data complexity, its varying performance indicates the distinct characteristics of these two datasets. 

Next, we examine the impact of different training configurations. We observe that the 3B model achieves a higher speedup ratio than the 0.5B model. This is because the attention mechanism constitutes a more severe computational bottleneck in the 3B model, amplifying the load imbalance issues. Consequently, optimizing the 3B model yields more efficiency benefits. Additionally, the 3B LoRA configuration exhibits speedups similar to the standard 3B model, demonstrating the versatility and robustness of \ours across different training paradigms.

\subsubsection{Ablation Study}
\label{sec:ablation}

To evaluate the impact of the key hyper-parameters in \ours, we conduct an ablation study on the sparsity threshold $p$, the execution anchor choice, and the micro-batch size using the Qwen2.5-3B model. We test the speedup ratio under a configuration of $\text{DP}=2, \text{PP}=4, \text{TP}=1, \text{SP}=1$ in this section. 

\begin{figure}[htbp]
    \centering
    \includegraphics[width=0.7\linewidth]{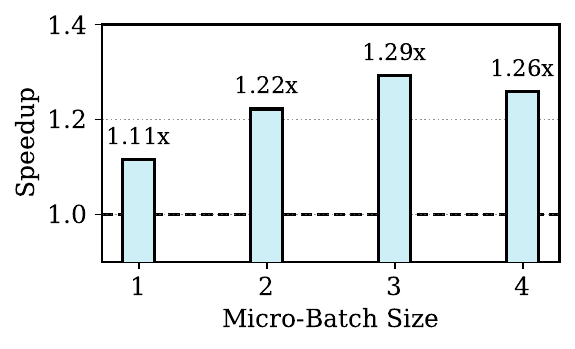}
    % \caption{Performance sensitivity across different micro-batch sizes.}
    \caption{Sensitivity of end-to-end speedup to different micro-batch sizes ($mbs=1$ to $4$, corresponding to global batch sizes $gbs=8$ to $32$). \ours consistently achieves clear speedups across all tested settings.}
    \label{fig:ablation_mbs}
\end{figure}

First, we study the impact of $p$ and anchor choice. With the anchor fixed to \cfgMean, increasing $p$ from $0.05$ to $0.30$ steadily improves the end-to-end speedup from $1.074\times$ to $1.451\times$, indicating that a looser threshold enables more aggressive optimization. 
Second, we evaluate the two extreme anchor choices, \cfgMin and \cfgMax, which align the workload to the predicted shortest and longest micro-batch latencies, respectively. 
The range of bars delineates the optimization potential within this dimension. 
Last, Figure~\ref{fig:ablation_mbs} further shows that \ours consistently achieves notable speedups across all tested micro-batch sizes, demonstrating its robustness to different batching sizes.

\begin{figure}[htbp]
    \centering
    \includegraphics[width=0.8\linewidth]{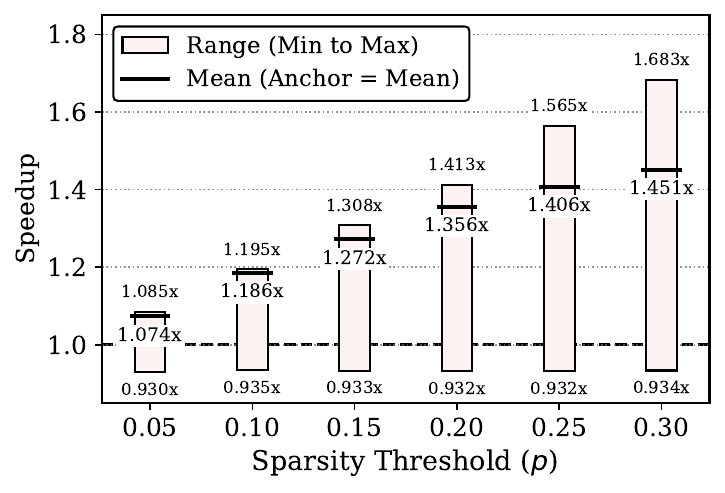}
    \caption{Sensitivity of end-to-end speedup with respect to the sparsity threshold $p$ and execution anchor choice. The upper end of each bar corresponds to \cfgMin, the lower end corresponds to \cfgMax, and the horizontal line denotes \cfgMean.}
    \label{fig:ablation_p_alpha}
\end{figure}

\subsubsection{Case Study}
\label{sec:case_study}

To further understand where the efficiency gains of \ours come from, we conduct a case study that jointly examines workload imbalance and overheads in iteration level. In particular, we compare \ours with a \textit{Length-Based Batching} (LBB) strategy, which uses sequence length as the batching weight and serves as a representative baseline for existing batching methods. In contrast, \ours uses latency prediction to guide workload partitioning. We conduct this study on Qwen2.5-3B with ChatQA2 under a single-node setting with $\text{DP}=2$, $\text{PP}=4$, micro-batch size $=2$, and global batch size $=16$.

\begin{table}[t]
\centering
\caption{LBB denotes a length-based batching strategy. Imbalance measures the average micro-batch-level imbalance degree (lower is better). Overheads report the additional cost introduced by \ours.}
\label{tab:imbalance_speedup_eval}
% 使用 \linewidth 确保表格完美自适应单栏宽度
\resizebox{\linewidth}{!}{
\begin{tabular}{lcccc}
\toprule
\textbf{Config.} & \textbf{Imbalance $\downarrow$} & \textbf{Iter. (ms)} & \textbf{Overheads (ms)} & \textbf{Speedup $\uparrow$} \\
\midrule
Baseline   & 1.81$\times$ & 4202.09 & 0.00 & 1.00$\times$ \\
+DST       & 1.73$\times$ & 3901.57 & 37.14 & 1.08$\times$ \\
+SAB       & 1.41$\times$ & 3470.73 & 1.81 & 1.21$\times$ \\
+LBB       & 1.58$\times$ & 3421.98 & 0.00 & 1.23$\times$ \\
+SAB+DST   & 1.34$\times$ & 3107.83 & 44.23 & \textbf{1.35$\times$} \\
+LBB+DST   & 1.41$\times$ & 3272.88 & 39.01 & 1.28$\times$ \\
\bottomrule
\end{tabular}
}
\end{table}

We quantify the workload imbalance in each iteration as
\begin{equation}
\label{eq:imbalance}
    Imbalance = \frac{\max(T_{mb,i})}{\frac{1}{N}\sum T_{mb,i}}
\end{equation}
where $T_{mb,i}$ denotes the average execution time of micro-batch $i$ across ranks. This metric reflects how unevenly the workload is distributed within an iteration. Although it is not directly proportional to the end-to-end iteration time, it provides a useful diagnostic signal for analyzing the straggler effect.

Table~\ref{tab:imbalance_speedup_eval} reveals three observations. First, the overheads of \ours remains small. SAB introduces negligible overhead, while the full SAB+DST pipeline adds only 44.23 ms, which is below 1.5\% of the total iteration time. 
Second, SAB achieves a slightly lower standalone speedup than LBB. We attribute this gap to the fact that SAB is designed to cooperate with the subsequent runtime adjustment in DST, and therefore performs a more aggressive data reorganization based on the predicted sparsity. When DST is disabled, this prediction cannot be fully realized at runtime, making sub-optimal efficiency.
Third, once combined with DST, SAB consistently outperforms LBB in both imbalance degree and end-to-end speedup, highlights the strong synergy between SAB and DST.

\subsection{Accuracy Evaluation}
\label{sec:accuracy_eval}

In this section, we evaluate whether \ours preserves model quality while improving system efficiency. We first analyze training loss to verify stable convergence, and then evaluate the fine-tuned checkpoints on downstream benchmarks covering general reasoning, long-context understanding, and exact retrieval.

% In this section, we evaluate the model accuracy of \ours, including the training loss and downstream benchmark performance. We conduct long context supervised training with Qwen-2.5-3B model using the ChatQA2 dataset. We employ MoBA as our base sparse attention algorithm with the block size = 256 and Top-K= 32. 

\begin{figure}[htbp!]
    \centering
    
    \includegraphics[width=1\linewidth]{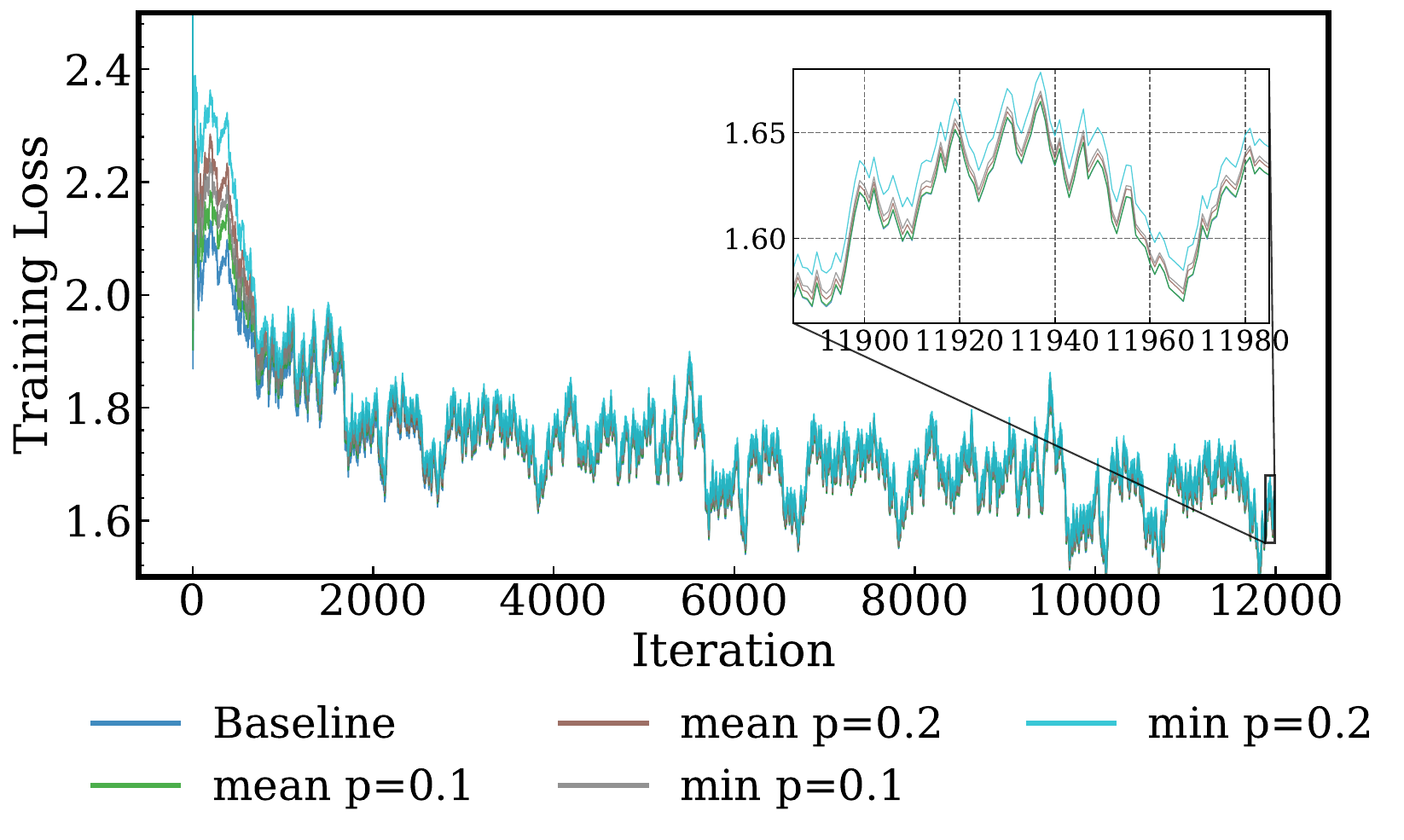}
    \caption{Training loss comparison between the MoBA baseline and \ours with different dynamic tuning configurations.}
    \label{fig:loss}
\end{figure}

\begin{table}[htbp!]
\centering
\caption{Average accuracy (\%) of Needle-in-a-Haystack (NIAH) tasks evaluated on Qwen-2.5-3B model. Results are averaged across the \textit{niah\_single\_1}, \textit{niah\_single\_2}, and \textit{niah\_single\_3} evaluations for each model.}
\label{tab:3b_eval_niah}
\small % 使用 \small 保证与其他表格字体大小一致
\begin{tabular}{lcccc}
\toprule
\textbf{Model} & \textbf{$\le$8K} & \textbf{16K} & \textbf{32K} & \textbf{64K} \\
\midrule
Baseline SFT & 100.00 & 100.00 & 100.00 & \textbf{97.60} \\
\midrule
\cfgMin[0.1] & 100.00 & 100.00 & 99.93 & 96.33 \\
\cfgMin[0.2] & 100.00 & 100.00 & 99.93 & 95.20 \\
\cfgMean[0.1] & 100.00 & 100.00 & 100.00 & 97.53 \\
\cfgMean[0.2] & 100.00 & 100.00 & 99.93 & 96.47 \\
\bottomrule
\end{tabular}
\end{table}

\subsubsection{Training Loss}
To validate the model convergence and accuracy, we conduct long-context fine-tuning on the Qwen-2.5-3B model using the ChatQA2 dataset. We employ MoBA with a block size of 256 and top-k 32 as our baseline. For \ours, we evaluate configurations with \cfgMin[0.1], \cfgMin[0.2], \cfgMean[0.1], and \cfgMean[0.2]. We exclude the \cfgMax configurations, as they severely degrade training efficiency without practical benefits.
% , which aligns with our findings in Section~\ref{sec:ablation}.

As illustrated in Figure~\ref{fig:loss}, the training loss curves of \ours closely track the baseline. Although minor deviations appear during the early training stages—likely stemming from the instability of transitioning from dense pre-training to sparse fine-tuning—the gap rapidly diminishes. Notably, the configuration with $p=0.2$ exhibits a slightly higher loss than $p=0.1$. This underscores the necessity of the performance-bound mechanism in our DST module, which effectively prevents excessive critical information dropping in sensitive sequences. Furthermore, setting the anchor \cfgMean tends to yield lower loss compared to targeting the minimum latency \cfgMin. This is consistent with the design intuition of bidirectional tuning in \ours.
% This aligns with our core design philosophy: reallocating attention budgets to reduce the sparsity of non-bottleneck micro-batch effectively provides a ``free accuracy'' improvement by utilizing system bubbles.

\begin{table}[t]
\centering
\caption{General benchmark results on Qwen2.5-3B. \textbf{Bold} indicates the best performance.}
\label{tab:3b_eval_gen}
\setlength{\tabcolsep}{2.8pt}
\renewcommand{\arraystretch}{0.95}
\begin{tabular}{lccccc}
\toprule
\textbf{Task} & \textbf{Base-SFT} & \textbf{\cfgMin[0.1]} & \textbf{\cfgMin[0.2]} & \textbf{\cfgMean[0.1]} & \textbf{\cfgMean[0.2]} \\ 
\midrule
ARC-Chal. & 0.4727 & 0.4735 & 0.4684 & \textbf{0.4744} & 0.4659 \\
ARC-Easy      & 0.7332 & \textbf{0.7348} & 0.7277 & 0.7315 & 0.7260 \\
BoolQ         & 0.7936 & 0.7994 & 0.7951 & 0.8003 & \textbf{0.8031} \\
HellaSwag     & 0.7356 & 0.7354 & 0.7355 & 0.7349 & \textbf{0.7366} \\
Lambada       & 0.6598 & 0.6583 & 0.6594 & \textbf{0.6604} & 0.6592 \\
Piqa          & 0.7884 & 0.7867 & 0.7894 & \textbf{0.7911} & 0.7900 \\
Winogrande    & \textbf{0.6875} & 0.6851 & 0.6835 & 0.6843 & 0.6859 \\ 
\midrule
\textbf{Average} & 0.6958 & 0.6962 & 0.6941 & \textbf{0.6967} & 0.6952 \\ 
\bottomrule
\end{tabular}
\end{table}

\subsubsection{Benchmark Evaluation}
To thoroughly evaluate the model capabilities, we assess the fine-tuned models on three benchmarks, representing distinct core abilities: long-context understanding (LongBench), exact retrieval (Needle-in-a-Haystack, NIAH) and standard zero-shot reasoning (General Benchmarks).

\textbf{Long-Context Understanding.} Table~\ref{tab:3b_eval_longbench} reports the category-level LongBench results, which serve as our primary downstream evaluation for long-context capability. Among all evaluated configurations, \cfgMean[0.1] achieves the best overall score (39.46), outperforming the Baseline SFT (39.28). In contrast, relaxing the threshold to $p=0.2$ generally causes slight degradation, indicating that overly aggressive tuning hurts model fidelity. We also observe that \cfgMean[0.1] provides a better overall balance than \cfgMin[0.1], which is consistent with the design intuition that bidirectional tuning can exploit non-critical-path computation for accuracy improvement. We further note that our fine-tuning data is QA-oriented (ChatQA2-Long-SFT), which helps to explain why the gains of \ours are more apparent on QA tasks.

% \begin{table}[t]
% \centering
% \caption{Category-level average results on LongBench for Qwen2.5-3B.}
% \label{tab:3b_eval_longbench}
% \small
% \setlength{\tabcolsep}{4pt}
% \renewcommand{\arraystretch}{0.95}
% \begin{tabular}{lccccc}
% \toprule
% \textbf{Category} & \textbf{Base-SFT} & \textbf{\cfgMin[0.1]} & \textbf{\cfgMin[0.2]} & \textbf{\cfgMean[0.1]} & \textbf{\cfgMean[0.2]} \\
% \midrule
% Single-Doc QA & 32.24 & \textbf{33.86} & 33.68 & 32.93 & 32.99 \\
% Multi-Doc QA  & 34.18 & 35.56 & 35.04 & \textbf{35.84} & 35.41 \\
% Summarization & \textbf{26.14} & 24.71 & 24.92 & 25.63 & 24.38 \\
% Retrieval     & \textbf{49.08} & 47.10 & 48.30 & 48.34 & 49.00 \\
% Code          & \textbf{67.41} & 66.44 & 66.52 & 66.54 & 66.24 \\
% \midrule
% \textbf{Overall} & 39.28 & 39.19 & 39.28 & \textbf{39.46} & 39.14 \\
% \bottomrule
% \end{tabular}
% \end{table}

\textbf{Exact Retrieval.} Table~\ref{tab:3b_eval_niah} shows that all configurations achieve perfect retrieval up to 16K context length. Even at 64K, \cfgMean[0.1] remains nearly on par with the Baseline SFT (97.53\% vs. 97.60\%), indicating that DST does not noticeably harm exact long-context retrieval.

\textbf{General Benchmarks.} On standard zero-shot reasoning tasks, \ours remains highly competitive with the baseline. In particular, \cfgMean[0.1] achieves the best average score (0.6967), slightly outperforming the Baseline SFT (0.6958), which suggests that the proposed dynamic tuning does not degrade general model capability.

Overall, the accuracy results show that \ours preserves stable training and does not compromise downstream capability. Among the evaluated configurations, \cfgMean[0.1] consistently provides the best overall balance, while $p=0.1$ serves as a necessary strict bound for maintaining model fidelity. These results are consistent with the design of DST, where such bidirectional sparsity tuning strategy achieves a dual benefit of system efficiency and model accuracy.

\subsection{Trade-off Analysis}
\label{sec:trade_off_analysis}

\begin{figure}[htbp!]
    \centering
    \includegraphics[width=0.8\linewidth]{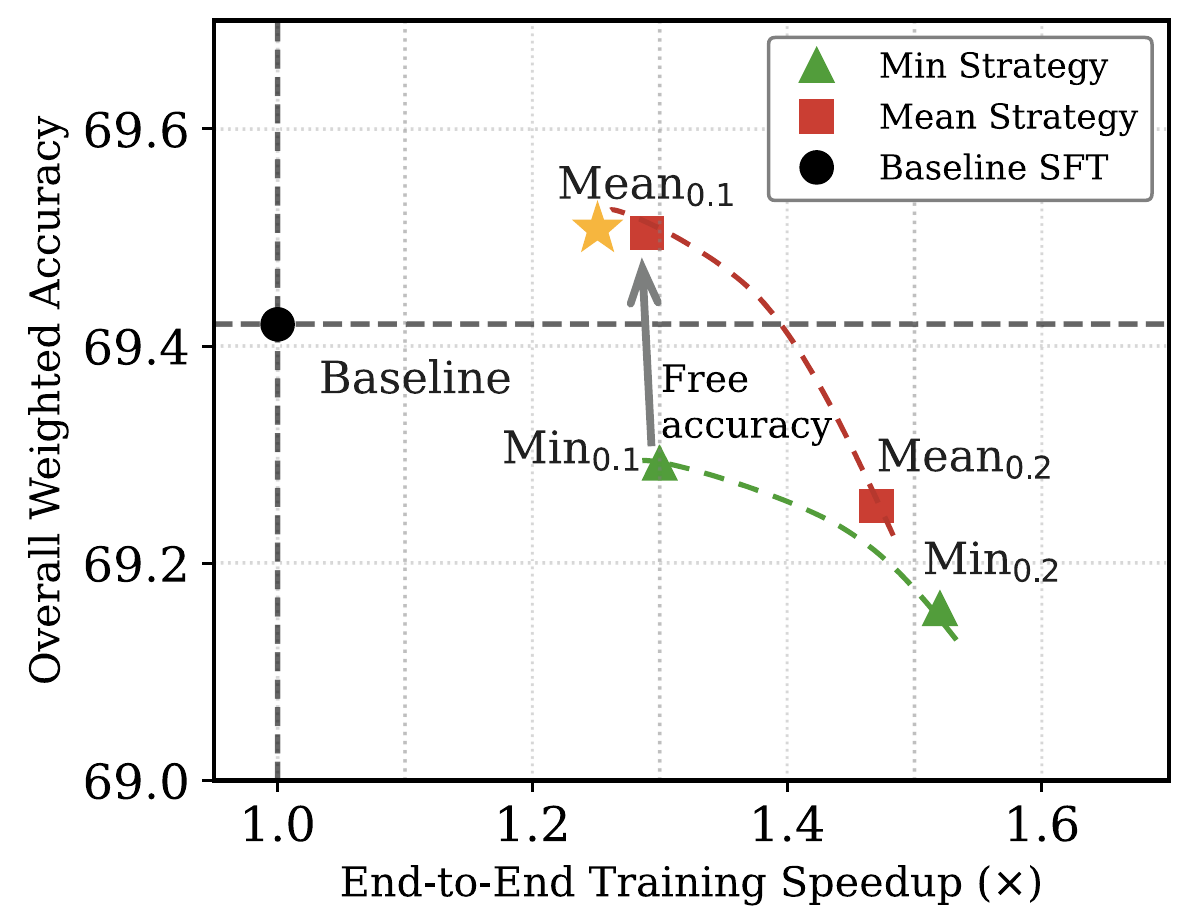}
    \caption{Trade-off between training speedup and downstream model performance under different configurations in \ours.}
    \label{fig:pareto_speed_vs_accuracy}
\end{figure}

The previous sections evaluate the efficiency and model quality of \ours separately. Here, we jointly examine these two aspects to understand the practical operating point of dynamic sparsity tuning and the inherent trade-off in \ours. 
To directly visualize this trade-off, we plot the relation between training speedup and model accuracy under different configurations in Figure~\ref{fig:pareto_speed_vs_accuracy}, where the accuracy is represented by the weighted average score across all evaluated benchmarks.

The resulting configurations form a clear Pareto frontier. In general, increasing the threshold $p$ moves the operating point toward higher speedup but lower model quality, confirming that $p$ directly governs how aggressive we trade accuracy for efficiency. Meanwhile, anchor point \cfgMean generally achieves better accuracy than the \cfgMin while delivering comparable system efficiency. Among all evaluated configurations, \cfgMean[0.1] lies at the most favorable point on this frontier, which improves both efficiency and accuracy. Therefore, we use \cfgMean[0.1] as the default practical setting in the efficiency evaluation of Section~\ref{sec:efficiency_eval}.

Overall, these results consolidate our claim that, through system-aware bidirectional sparsity tuning, \ours can achieve a dual benefit in both system efficiency and model accuracy.

\section{Related Work}

\textbf{System Optimization in Long-Context Training.} 
To enhance the training efficiency of long-context LLMs, existing systems primarily rely on hybrid parallelism strategies \cite{megatron_v1_tp_2020, megatron_v2_3d_par_interleave1f1b_2021, megatron_v3_sp_selective_recompute_2023, megatron_deepspeed_2022}, which combine data, pipeline, tensor, and sequence parallelism. However, the highly heterogeneous sequence length distribution in real-world long-context data inherently leads to severe load imbalance across workers. To mitigate this issue, various batching and scheduling techniques have been proposed \cite{wang2025wlb, bai-etal-2024-longalign, xu2025skrull, chunkflow}. For example, LongAlign adopts sorted batching to group sequences by length within a global batch \cite{bai-etal-2024-longalign}, 
while WLB-LLM employs a heuristic packing approach that adaptively delays the execution of long sequences \cite{wang2025wlb}.
Despite their effectiveness in improving system throughput, these methods mainly optimize data organization and may compromise data randomness, potentially raising the risk of changing the convergence behavior. Furthermore, existing works are largely designed for dense attention training and fall short in sparse scenarios. As a result, they do not address the additional workload uncertainty introduced by dynamic sparse attention.

\textbf{Sparse Attention Training.} 
Sparse attention has become a promising direction for reducing the quadratic cost of long-context modeling. Early work mainly focused on training-free sparse methods for inference acceleration \cite{zhang2023h2o, Quest}, while recent efforts have shifted toward trainable sparse attention mechanisms. Representative methods such as NSA \cite{yuan-etal-2025-nsa}, MoBA \cite{lu2025mobamixtureblockattention}, and DSA \cite{deepseekai2025deepseekv32pushingfrontieropen} accelerate training by selecting critical tokens or blocks with Top-K-style mechanisms. However, these methods typically use a fixed attention budget across sequences and layers, overlooking the substantial heterogeneity in sparsity sensitivity during training. More recent work has begun to explore dynamic sparse attention. For instance, Twilight \cite{lin2025twilight} introduces hierarchical Top-P selection for adaptive budget allocation, while MTraining \cite{li2025mtraining} combines dynamic sparsity guided by a vertical-slash structure with context-parallel execution optimization. 
Nevertheless, these methods primarily treat dynamic sparsity as an algorithmic mechanism, while \ours further makes sparsity control system-aware by coupling runtime tuning with profiling-based latency modeling and workload balancing.

% Nevertheless, these methods primarily treat dynamic sparsity as an algorithmic optimization, neglecting the crucial system-level insights. In contrast, \ours explicitly couples runtime sparsity tuning with latency prediction and balanced batching strategy, enabling joint optimization of model accuracy and distributed training efficiency.

\section{Conclusion}

In this paper, we propose \ours, a novel algorithm-system co-design framework for long-context sparse training. It addresses a central challenge in these scenarios, where sequence-length heterogeneity, sparsity heterogeneity, and distributed workload imbalance are tightly coupled during training. To tackle this problem, we develop workload-aware dynamic sparsity tuning for fine-grained runtime balancing, a profiling-based latency predictor for practical workload estimation, and sparsity-aware batching for coarse-grained initialization. Experiments on real-world long-context datasets and evaluations on downstream tasks show that \ours improves end-to-end training efficiency by up to 1.33$\times$ while still improving the model's long-context capability. These results suggest that dynamic sparsity should not only be viewed as an algorithmic feature, but can also serve as an effective approach for system-level load balance in distributed training. We hope \ours can motivate future work on jointly optimizing model capabilities and system efficiency for long-context LLM training.

\bibliographystyle{IEEEtran}
\bibliography{ref/sc_cite,ref/framework,ref/mainstream_llm}
% \printbibliography

\end{document}